\documentclass[10pt,twocolumn,letterpaper]{article}

\usepackage{cvpr}
\usepackage{times}
\usepackage{epsfig}
\usepackage{graphicx}
\usepackage{amsmath}
\usepackage{amssymb}

\usepackage{amssymb} 
\usepackage{booktabs} 
\usepackage{subcaption} 
\usepackage{verbatim} 
\usepackage{caption} 
\usepackage{multirow} 
\usepackage{gensymb} 
\usepackage[dvipsnames]{xcolor} 
\DeclareMathOperator*{\argmax}{arg\,max} 

\usepackage[pagebackref=true,breaklinks=true,letterpaper=true,colorlinks,bookmarks=false]{hyperref}

\cvprfinalcopy 

\def\cvprPaperID{329} 

\ifcvprfinal\fi
\begin{document}

\title{Pairwise Decomposition of Image Sequences for Active Multi-View Recognition}

\author{Edward Johns, Stefan Leutenegger and Andrew J. Davison\\\normalsize{Dyson Robotics Laboratory at Imperial College, Department of Computing, Imperial College London, UK}
}

\maketitle
\thispagestyle{empty}

\begin{abstract}
A multi-view image sequence provides a much richer capacity for object recognition than from a single image. However, most existing solutions to multi-view recognition typically adopt hand-crafted, model-based geometric methods, which do not readily embrace recent trends in deep learning. We propose to bring Convolutional Neural Networks to generic multi-view recognition, by decomposing an image sequence into a set of image pairs, classifying each pair independently, and then learning an object classifier by weighting the contribution of each pair. This allows for recognition over arbitrary camera trajectories, without requiring explicit training over the potentially infinite number of camera paths and lengths. Building these pairwise relationships then naturally extends to the next-best-view problem in an active recognition framework. To achieve this, we train a second Convolutional Neural Network to map directly from an observed image to next viewpoint. Finally, we incorporate this into a trajectory optimisation task, whereby the best recognition confidence is sought for a given trajectory length. We present state-of-the-art results in both guided and unguided multi-view recognition on the ModelNet dataset, and show how our method can be used with depth images, greyscale images, or both.
\end{abstract}

\begin{figure}[t]
    \centering
    \includegraphics[width=\linewidth]{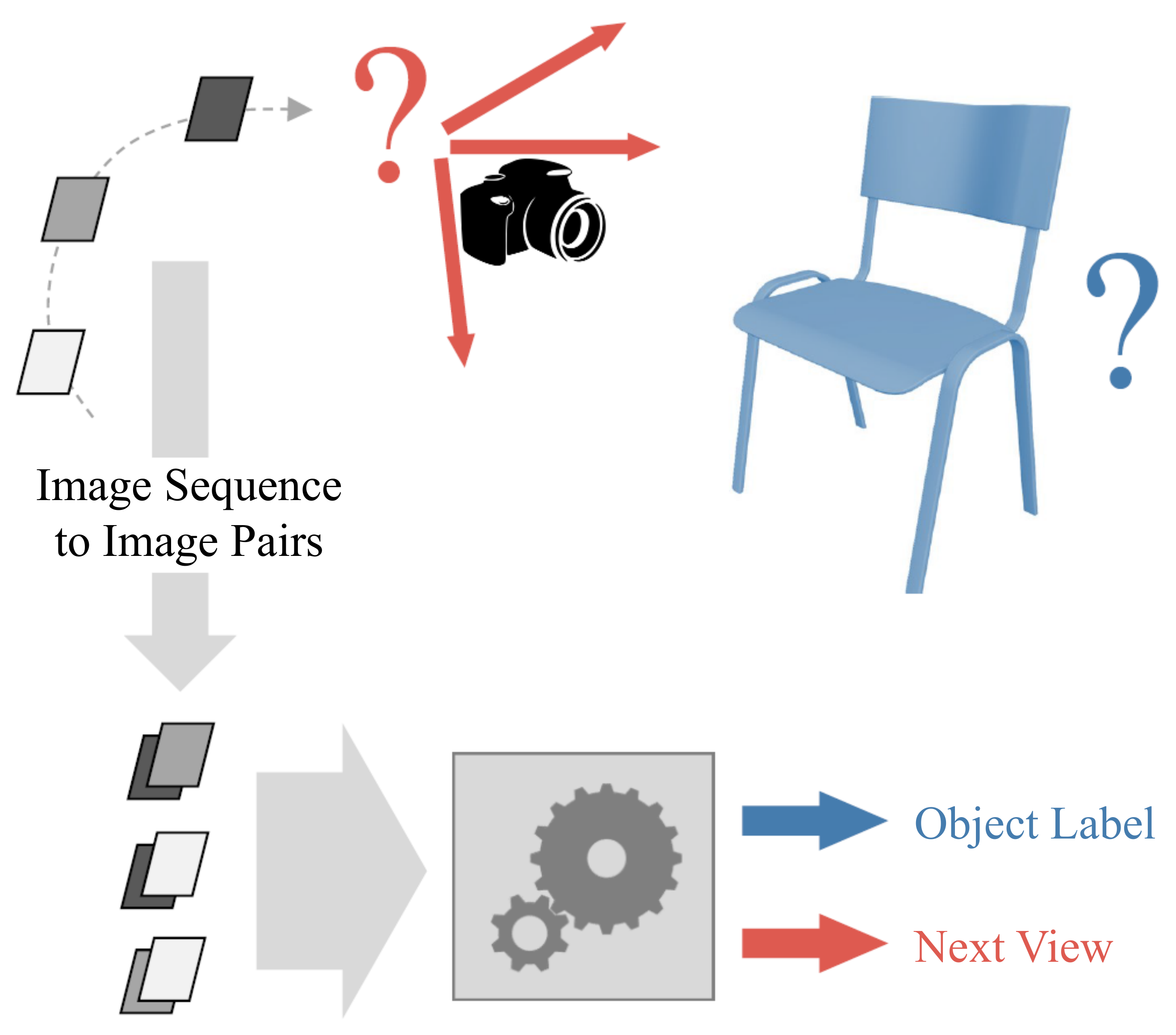}
    \caption{We propose a method for multi-view object recognition, by decomposing an image sequence into a set of image pairs. Training on these pairs then allows for recognition and trajectory planning, without the need to train directly over the infinite possible number of camera paths that may exist. }
    \label{fig:overview}
\end{figure}

\section{Introduction}

Consider the scenario in Figure \ref{fig:overview}. What trajectory should the camera move around the object in order to achieve the highest recognition confidence in a given time? For practical tasks, recognition from a \emph{multi-view} image sequence is a more realistic setting than the single-image recognition tasks typically addressed in computer vision, and controlling a camera \emph{actively} for efficient recognition has great significance in real-world applications, where time or power constraints become realities. For example, a robot rotating an object before its eyes, or a mobile robot semantically mapping a room, benefit from efficient solutions.

Traditionally, multi-view object recognition has been achieved by building up compositions of hand-crafted features shared across viewpoints, and finding correspondences between a test image and the learned models \cite{Lowe:CVPR2001, Thomas:etal:CVPR2006, Pepik:etal:2015}. However, recent trends in Convolutional Neural Networks (CNNs) \cite{Krizhevsky:etal:NIPS2012, Simonyan:Zisserman:ICLR2015} have seen attention in single-view object recognition move away from these explicit, hand-modelled, geometric solutions, and towards end-to-end learning ideologies which inject fewer assumptions into the learned object models. Recently, the introduction of the ModelNet dataset of 3D CAD meshes \cite{Wu:etal:CVPR2015} provided data of sufficient magnitude for training deep networks with images covering the full sphere of viewpoints over an object, enabling view synthesis without the need for laborious manual labelling of each image \cite{Johns:etal:CVPR2015}. It was subsequently shown that rendering these meshes as synthetic greyscale images, and classifying objects in a view-based manner with a CNN architecture acting over a fixed trajectory, achieved state-of-the-art results for multi-view recognition \cite{Su:etal:ICCV2015}. However, extending this to generalised recognition over trajectories of arbitrary paths and lengths is not readily adopted by traditional CNN architectures, due to the need for fixed-length input data.

\subsection{CNNs for Generalised Multi-View Recognition}

One solution to multi-view recognition with CNNs would be to simply concatenate all observed images into a single input to a network. However, this would require intractable training due to the large size of each input, but more importantly, due to the need to train over every possible path of all possible lengths, which is of potentially infinite scale. We propose to address this by relaxing the joint model over images and decomposing an image sequence into a set of pairs, one for every pair of images across the sequence. Pairwise representations of full distributions have been popular in computer vision for learning distributions of local features \cite{Johns:Yang:ECCV2014} and parts \cite{Felzenszwalb:etal:PAMI2010}, and we migrate this idea from the image space domain to the temporal domain. Given this decomposition, a CNN is then trained on a fixed-length input consisting of the image pair, together with the relative pose between the associated viewpoints. To achieve classification of the full sequence, an ensemble framework is adopted, with weighting to increase the contribution of those image pairs which cover a more informative set of poses.

The problem then shifts to active recognition, with the aim of determining along which trajectory the camera should move, in order to achieve the best recognition accuracy in a given number of images. This is often presented as a Next-Best-View (NBV) prediction, where the mutual information is determined between the class probability distribution and each potential next view. However, this typically requires learning a generative model of the object and synthesising new views as an intermediate step. We propose to learn NBV prediction with a more powerful discriminative model, training a second CNN to map directly from an observed image to the rotation angle over which the camera should subsequently move.

Finally, we extend our NBV prediction to a full trajectory-optimisation framework, where we consider all possible images that can acquired along a trajectory as contributions, rather than simply following a sequence of NBV images as is often employed. To achieve this, we train a third CNN in a similar manner to the above NBV CNN, but training for regression to a recognition confidence score for all possible next viewpoints, rather then classification for the overall best viewpoint. As the image sequence evolves, all unvisited viewpoints accumulate scores based on the newly-observed images, and the optimum trajectory is chosen as the one which maximises the summation of these scores.

\subsection{Contributions}

In this paper, we present three key technical contributions all based on powerful CNN learning:

\begin{enumerate}
\item Multi-view object recognition over arbitrary camera trajectories by training only on image pairs,
\item Discriminatively-trained Next-Best-View prediction directly from an input image to the next viewpoint,
\item Trajectory optimisation by considering the impact of all observable images along the sequence.
\end{enumerate}

All three contributions achieve state-of-the-art results in their respective benchmarks on the ModelNet dataset \cite{Wu:etal:CVPR2015}.

\section{Related Work}

\paragraph{View-Based Multi-View Recognition}
In its simplest form, the view-based approach aims to add viewpoint tolerance to a 2D image of an object, such as with viewpoint-invariant local descriptors \cite{Mikolajczyk:Schimd:PAMI2005, Philbin:etal:CVPR2007} or deformation-tolerant global descriptors \cite{Dalal:Triggs:CVPR2005}. Given training images across multiple viewpoints, a more stable set of features can be found by tracking those which are shared across multiple views and clustering images accordingly \cite{Lowe:CVPR2001}, or by learning their relative 2D displacements as the viewpoint changes, both with hard constraints for rigid bodies \cite{Johns:Yang:ICCV2011, Johns:Yang:IJCV2014} and flexible constraints for deformable bodies \cite{Fergus:etal:CVPR2003, Felzenszwalb:etal:PAMI2010}. To add further fidelity to the true underlying object geometry, these 2D image elements can also be embedded within an implicit 3D model \cite{Thomas:etal:CVPR2006, Liebelt:Schmid:CVPR2010, Pepik:etal:2015}. If multiple views are available at testing, images can be combined and treated as a single, larger image \cite{Pless:CVPR2003}, an approach which can also be addressed in two stages, by processing the individual images first to reduce the search space \cite{Collet:Srinivasa:ICRA2010}.

Recently, CNN architectures have been extended to allow for recognition from image sequences using a single network, by max pooling across all viewpoints \cite{Su:etal:ICCV2015}, or by unwrapping an object shape into a panorama and max pooling across each row \cite{Shi:etal:SPL2015}. However, both these methods assume that a fixed-length image sequence is provided during both training and testing, and hence are unsuitable for generalised multi-view recognition.

\paragraph{Shape-Based Multi-View Recognition}
Rather than modelling an object as a set of views with 2D features, an explicit 3D shape can be learned from reconstruction \cite{Vicente:etal:CVPR2014} or provided by CAD models \cite{Wu:etal:CVPR2015}, and subsequently matched to from depth images \cite{Gupta:etal:ECCV2014}, 3D reconstructions \cite{Bai:etal:CVPR2016}, or partial reconstructions with shape completion \cite{Firman:etal:CVPR2016, Wu:etal:CVPR2015}. Shape descriptors include distributions of local surface properties \cite{Horn:IEEE1984, Osada:etal:ACMTOG2002}, spherical harmonic functions over voxel grids \cite{Kazhdan:etal:SGP2003}, and 3D local invariant features \cite{Knopp:etal:ECCV2010}. Recently, CNNs have been applied to 3D shapes by representing them as 3D occupancy grids, and building generative \cite{Wu:etal:CVPR2015} or discriminative \cite{Maturana:Schererl:IROS2015} networks.

As of now however, CNNs with 2D view-based methods \cite{Su:etal:ICCV2015} have outperformed their counterpart 3D voxel-based methods \cite{Wu:etal:CVPR2015, Maturana:Schererl:IROS2015}, and we therefore adopt the 2D approach in our work. However, it is not yet clear whether this greater performance arises from the superior abundance of 2D image data for pre-training deep networks, or the naturally more efficient representation of 2D than 3D in standard CNN architectures.

\paragraph{Active Recognition}
Methods for active recognition typically learn a generative model of the object, predict the object appearance from unvisited viewpoints, and select views based on a measure of entropy reduction. \cite{Wu:etal:ICRA2015} modelled objects as a 3D cloud of SIFT features, moving the camera to the view which would reveal the greatest number of features which have not yet been observed. A similar method was proposed in \cite{Charrow:etal:RSS2015} for guided mapping and robot navigation. The incorporation of active recognition into a Random Forests framework was presented in \cite{Doumanoglou:etal::ECCV2014}, whereby each decision tree encodes both object classification and viewpoint selection. Recently, the ShapeNets framework of \cite{Wu:etal:CVPR2015} proposed to model objects as a voxel grid, and learn a generative model based on Convolutional Deep Belief Networks to allow for view synthesis from unseen viewpoints.

However, these methods do not take into account the images acquired along a sequence towards the chosen next view. In \cite{Kraininl:etal:ICRA2011}, this was incorporated during active object reconstruction by visiting a sequence of actively-selected views, but reconstructing the object based on the entire image sequence that is observed between the views. For recognition, Partially Observable Markov Decision Processes (POMDPs) \cite{Eigenberger:Scharinger:IROS2010} have seen success in optimising a trajectory for a particular task, although these require generative modelling rather than direct discriminative learning as we propose in our method. Finally, recurrent CNNs have recently been shown to be effective for active recognition from image sequences \cite{Ba:etal:ICLR2015}, and we believe that this approach has exciting future potential.

\section{Multi-View Object Recognition}

\subsection{Dataset}

We train and test our proposed methods on the ModelNet dataset of 3D CAD meshes \cite{Wu:etal:CVPR2015}, which provides multi-view training data of sufficient scale for training deep networks. As in \cite{Wu:etal:CVPR2015, Su:etal:ICCV2015}, we discretise viewpoints into distinct steps, but whereas in these works rotations are constrained to being around the gravity vector, we relax this and allow rotations around the object's full viewing sphere to enable recognition from arbitrary camera trajectories. The camera pose is defined in spherical coordinates $\{r, \theta, \phi\}$, where $r$ is fixed and $\theta$ and $\phi$ are divided into 30$^{\circ}$ steps, and the camera is pointed towards the object's centroid. Camera paths then visit viewpoints along the viewing sphere with $\theta$ and $\phi$ either decreasing or increasing by one step, or remaining the same, between viewpoints. For every viewpoint, we render a greyscale image of the object object as with \cite{Su:etal:ICCV2015}, together with a depth image for dual-modality imaging.

For comparisons with \cite{Wu:etal:CVPR2015, Su:etal:ICCV2015}, we also assume each object to be aligned in its canonical orientation as defined in the ModelNet dataset, although augmenting the training data by rotating models as necessary would allow for relaxation of this prior assumption. As with these works, we also assume the pose of the camera to be known with respect to the object's viewing sphere, whereas in practice this would be achieved by visual tracking or reconstruction, or by use of robot kinematics or other external sensors. Training and testing models are provided as CAD meshes and free from occlusion or clutter, although in practice a detection and segmentation task would precede our pipeline.

\subsection{Pairwise Learning}

Our proposed multi-view object recognition method requires computing the probability over class labels given a sequence of $M$ views. To allow for flexibility of camera trajectories over all possible paths and lengths, we decompose a sequence into a set of $N = \frac{M(M-1)}{2}$ \emph{view pairs}, 
denoted $w_1 ... w_N$. Here, every new view acquired along a sequence forms a new view pair with all existing views in the sequence, and the task then becomes to compute a recognition score over all classes, $f(\mathbf{y} | w_1 ... w_N)$. 

The data for each view pair $w_i$ is composed of three elements: the image $x_i^1$ from the first view, the image $x_i^2$ from the second view, and the relative camera pose $\psi_i$ between the two views, such that $w_i = \{x_i^1, x_i^2, \psi_i\}$. For object recognition from a view sequence, we classify each view pair independently, and then weight the contribution from each, discussed further in Section \ref{sec:weights}. In this way, each view pair $w_i$ is processed with a \emph{weak} classifier, with an associated weight $\lambda_i$, and a \emph{strong} classifier then computes the weighted average of these for a final distribution of scores over class labels:

\begin{equation}
f(\mathbf{y} | w_1 ... w_N) = \sum_{i=1}^{i=N} \lambda_i \text{ } p(\mathbf{y} | w_i)
\label{eq:boosting}
~.
\end{equation}

\begin{figure}[h]
    \centering
    \includegraphics[width=\linewidth]{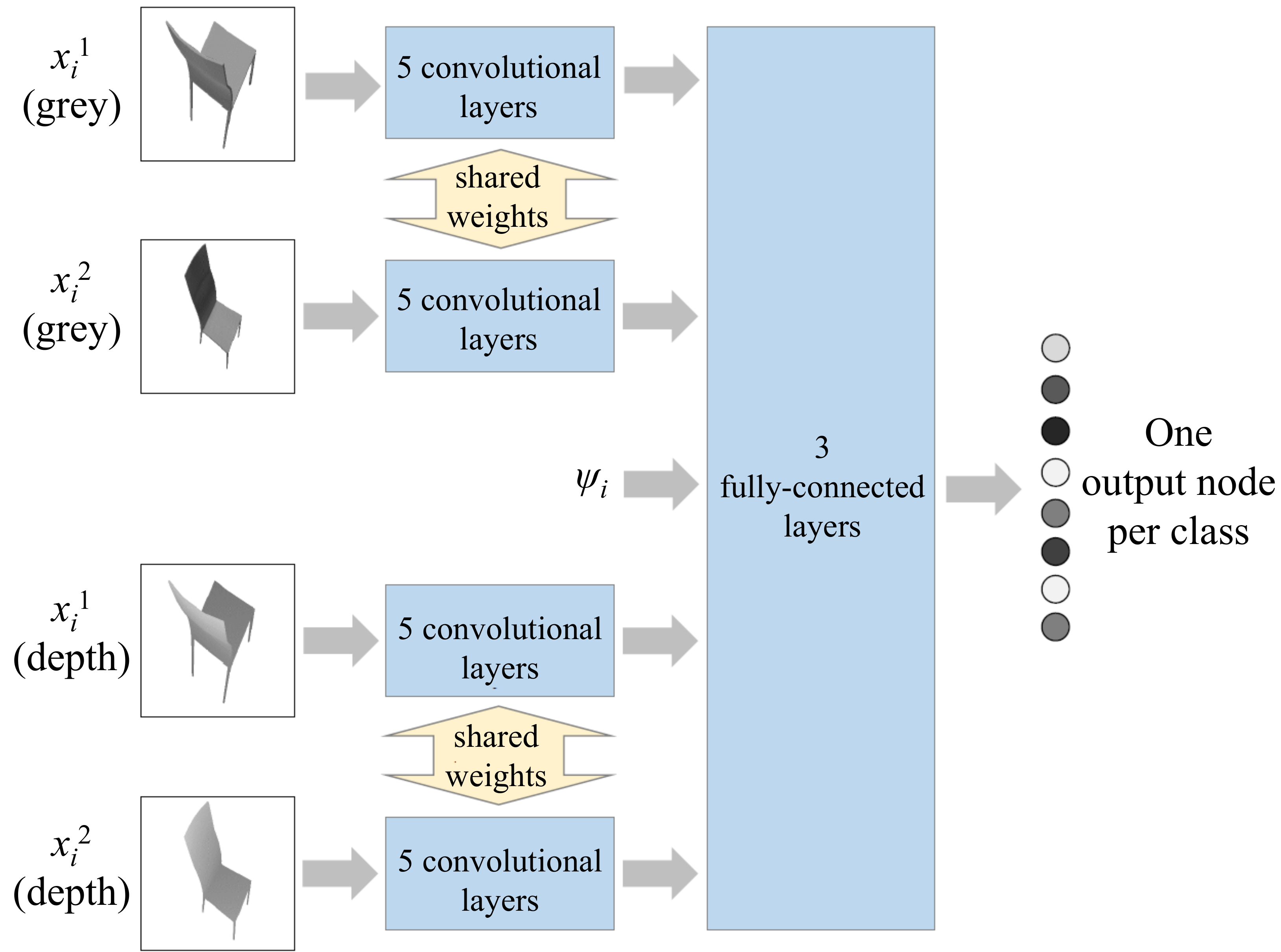}
    \caption{Our CNN-1 architecture for classification of a view pair, for use with greyscale images, depth images, or both.}
    \label{fig:cnn_1}
\end{figure}

To compute the class probability distribution $p(\mathbf{y} | w_i)$ for each view pair, we designed a CNN architecture, denoted \emph{CNN-1} (see Figure \ref{fig:cnn_1}), to predict an object class based on the provided view pair. This architecture was inspired by the Siamese CNN \cite{Chopra:etal:CVPR2005}, which consists of two CNN's running in parallel, each taking in one image from the pair, and with weights shared across both networks. Whilst this architecture is typically used to enforce similarity or dissimilarity between the outputs of the two networks, we use it to reduce the number of parameters to be learned, rather than concatenating the two images into a single input vector and training a larger network. Not only does this allow for efficient training, but also a fair comparison with the state-of-the-art \cite{Su:etal:ICCV2015} because we are not adding additional capacity to the network, as this has been shown to dramatically improve classification performance \cite{Simonyan:Zisserman:ICLR2015}.

For the convolutional layers of CNN-1, we follow \cite{Su:etal:ICCV2015} and adopt the VGG-M network \cite{Chatfield:etal:BMVC2014} with five convolutional layers and three fully-connected layers. The final convolutional layers from the two images are concatenated, together with a vector using one-hot encoding to represent the relative camera pose between the two views. Depending on the available imaging modality, the framework can be used with greyscale or depth images, or both. When both are used, we also concatenate the outputs of the convolutional layers for both modalities. Finally, three fully-connected layers are added after this concatenation, with classification loss computed using softmax and cross-entropy.

\subsection{Learning the weights}

\label{sec:weights}

Together with the two images, CNN-1 receives the relative pose of the view pair, and we use this to condition the classifier on how confident its output is likely to be. For example, a pair of images captured from two viewpoints at opposite ends of the viewing sphere may be more likely to be classified correctly than images from two adjacent viewpoints, because the former observes a greater coverage of the object and hence reveals more informative data upon which to make a classification decision.

We use cross entropy to measure the classification confidence for each relative pose, which computes the similarity between the ground truth distribution and the network's output distribution, a richer indicator than simply the classification error. For each relative pose $\psi_j$ in the discretised viewing sphere, the weight $\lambda_j$ is learned by averaging the cross entropy over all training image pairs whose viewpoints are separated by $\psi_j$. Then, all viewpoints in the sequence are weighted accordingly in Equation \ref{eq:boosting} to give greater importance to those view pairs which are likely to be classified correctly. Although the output class distribution of each pair already implicitly exhibits a measure of confidence based on the entropy, this additional weighting acts as a regularisation by injecting prior knowledge of how easily different viewpoint pairings can be separated, independently of the image content.

\section{Active Object Recognition}

\subsection{Next-Best-View Prediction}

\begin{figure}[h]
    \centering
    \includegraphics[width=\linewidth]{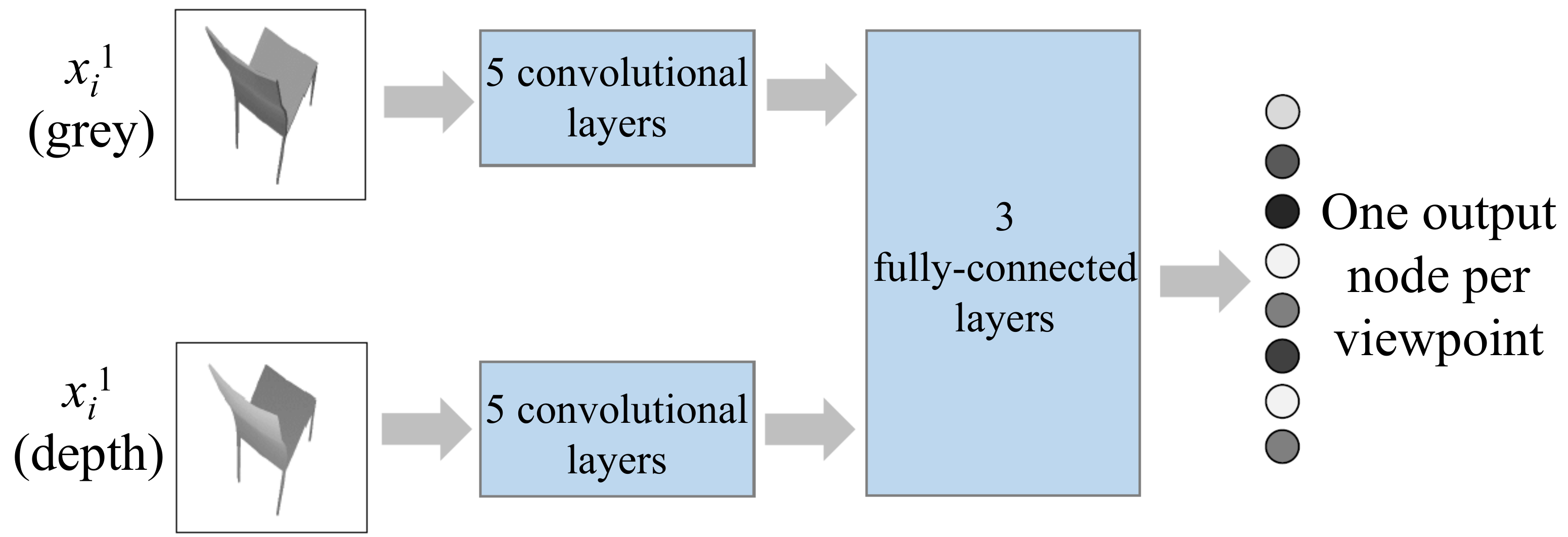}
    \caption{Our CNN-2 architecture for classifying the next-best-view given an input view, based on greyscale images, depth images, or both. This is also our CNN-3 architecture, where the output is a regression to predicted cross entropies over all viewpoints.}
    \label{fig:cnn_2}
\end{figure}

Given one view of an object, predicting the next view to move the camera to enables an active approach to recognition, by maximising the classification accuracy over a given number of views. Previous works \cite{Wu:etal:CVPR2015} typically address this by building a generative model of the object, predicting the observable image content from all other viewpoints, and choosing the view which, if observed, would reduce the class distribution entropy the most. We propose to solve this in a discriminative end-to-end manner, by training a second CNN, denoted \emph{CNN-2} (see Figure \ref{fig:cnn_2}), which directly outputs the best viewpoint to move to for any given input image. In this way, NBV prediction is trained in a direct and discriminative manner, with end-to-end learning which bypasses the intermediate step of generative prediction.

As with CNN-1, this network is based on the VGG-M network \cite{Chatfield:etal:BMVC2014}, with 5 convolutional and 3 fully-connected layers. However, rather than outputting a distribution over class labels as with CNN-1, the final layer consists of one node for every relative pose along the viewing sphere. To train CNN-2, every training image $x_k$ is paired with all other images from that same object, and the view pair is processed with CNN-1 to give a class distribution. Then, the view pair is chosen which yields the highest output for the ground truth class. The relative pose associated with this view pair, together with image $x_k$, then forms a training pair for CNN-2. During testing, a single view is passed through CNN-2, and the output node with the highest value determines the relative pose for the camera's next movement. A series of NBV movements can then be created by iterating this procedure sequentially.

\subsection{Trajectory Optimisation}
\label{sec:traj_opt}

Although this NBV prediction offers a simple solution to active recognition, it does not consider the images that could be acquired whilst the camera is moving towards the next view, and hence is not a globally-optimum solution. Rather than traversing a sequence of NBVs, recognition efficiency can be maximised by following a path which benefits from the contribution of \emph{all} observable images over the trajectory. Our proposed heuristic is that the optimum trajectory is one whose summation of predicted cross entropies, over all view pairs in the sequence, is smaller than for all other possible trajectories. In this way, we aim to contribute a high classification confidence to Equation \ref{eq:boosting} for every view pair, rather than only for the view pair formed from the first and last image, as is the case with CNN-2.

This is achieved as follows. For the current trajectory, let us denote the sequence of observed views as the set $v \in \mathcal{V}$. We then maintain a distribution $g(\boldsymbol{u})$ over all unobserved views $u \in \mathcal{U}$ on the viewing sphere, where $g_u$ indicates the cost of visiting view $u$ in the current trajectory. $g_u$ is defined as the sum of predicted cross entropies based on CNN-1, for all view pairs formed between $u$ and the set $\mathcal{V}$. As each new view is observed and added to the sequence, $g(\boldsymbol{u})$ is updated to reflect the scores based on the newly formed view pairs. The cost for each unobserved view is therefore:

\begin{equation}
 g_u = \sum_{v \in \mathcal{V}}h(u, v)
~.
\end{equation}
Here, $h(u, v)$ is the predicted cross entropy for classification of a view pair consisting of the unobserved view $u$, and the observed view $v$. To compute this value, we train a third CNN, denoted \emph{CNN-3} (see Figure \ref{fig:cnn_2}), which is identical to CNN-2 except that it is trained for regression to a cross entropy value, rather than being trained for classification to the next-best-view. CNN-3 maps a single input image to a distribution of predicted cross entropies over all viewpoints on the viewing sphere. Training pairs for this are generated by taking each training image, forming view pairs with all other views of that same object, and then computing the classification cross entropy with CNN-1. CNN-3 is then trained to minimise the L2 distance between the ground truth cross entropies and the predicted cross entropies.

We now define $\mathcal{U}^+$ as the set of 8 viewpoints adjacent to the camera's current viewpoint ($\Delta\theta = -30^{\circ}$, $0$, or $+30^{\circ}$, $\Delta\phi = -30^{\circ}$, $0$, or $+30^{\circ}$), representing all the positions which the camera can move to in its next step. For each viewpoint $u \in \mathcal{U}^+$, we compute the set of trajectories $\mathcal{T}_u$ over which the camera could subsequently traverse, if it were to make its next move to $u$. These are found by a simple undirected graph search, for a given final trajectory length. We then assign a score $s(t)$ to each trajectory $t \in \mathcal{T}_u$, by summing up the scores in the trajectory's set of unobserved views $\mathcal{U}_t$:

\begin{equation}
s(t) = \sum_{u \in \mathcal{U}_t} g_u
\end{equation}

The optimum next view $u^*$ is then chosen as the one whose best trajectory has the highest score over all of the available next views:
\begin{equation}
u^* = \argmax_{u \in \mathcal{U}^+} \text{    } \max_{t \in \mathcal{T}_u} \text{    } s(t)
~.
\end{equation}

In this way, at every step along the trajectory, the best decision is taken for the next view given the available information. As the camera follows this guided trajectory, the scores assigned to these unobserved views will change, attracting the camera towards those viewpoints which, if visited, are likely to yield a high classification confidence when processed with CNN-1.

\section{Experiments}

We evaluated our method on the ModelNet dataset \cite{Wu:etal:CVPR2015}, which consists of 3D CAD meshes from everyday objects over a range of scales. For our experiments, two subsets were used as in \cite{Wu:etal:CVPR2015}: ModelNet10, containing 10 object categories with 4,905 unique objects, and ModelNet40, containing 40 object categories and 12,311 unique objects, both with a testing-training split. ModelNet is the only available dataset at this time with sufficient large-scale multi-view coverage of objects for training or testing our networks, and hence as with \cite{Wu:etal:CVPR2015, Su:etal:ICCV2015}, real-world experiments were not possible.

Training of the three networks was then carried out by rendering images of each model from all viewpoints on the discretised viewing sphere, and forming the full set of image pairs. Rendering was performed under perspective projection, with objects scaled uniformly to fit the viewing window, to yield images of 512-by-512 pixels for both greyscale and depth images. For rendering the greyscale images, Phong shading \cite{Phong:CACM75} was used, and pre-training conducted with the ImageNet 1$k$ dataset \cite{Deng:etal:CVPR2009} as with \cite{Su:etal:ICCV2015}. Networks trained for the ModelNet10 dataset were pre-trained on ModelNet40. During testing, unless otherwise specified, every object was tested once per viewpoint, with the trajectory commencing at that viewpoint, and then the classification accuracy for that object was the average over all these trajectories.

\subsection{Pairwise Recognition}

First, we explored four different implementations of our multi-view recognition method, with both greyscale and depth images as input. We evaluated the performance based on two parameters: the weighting system used in Equation \ref{eq:boosting}, and the way in which view pairs are formed in a view sequence. Table \ref{tab:implementations} shows recognition results for random trajectories of 6 views, with and without learned weights for each view pair, such that without the weighting, all view pairs contribute equally to the final class distribution. Then, for selection of \emph{All} pairs, $\frac{M(M-1)}{2}$ pairs were chosen for a sequence of length $M$, such that every possible pair was used during recognition. For selection of \emph{Best} pairs, the top $M$ pairs with the greatest weight $\lambda$ were chosen, such that the number of pairwise classifications was linear rather than combinatorial with the sequence length. Results show the positive effect of the weighting and the inclusion of all pairs in the sequence, with the larger number of pairs offering slightly more benefit than the inclusion of weighting. For the remaining implementations of our method, weighting was used together with the full set of view pairs.

\begin{table}
\footnotesize
\centering
  \begin{tabular}{cccc}  
    \toprule
    Pair & \multirow{2}{*}{Weighted ?} & \multirow{2}{*}{ModelNet10} & \multirow{2}{*}{ModelNet40} \\
    Selection \\
    \midrule
    Best & No & 90.1 & 88.2 \\
    Best & Yes & 90.6 & 88.8 \\
    All & No & 91.2 & 89.0 \\
    All & Yes & \textbf{91.9} & \textbf{89.5} \\
    \bottomrule
  \end{tabular}
  \caption{Results for different implementations of our method over a sequence length of 6 views, for multi-view classification with Equation \ref{eq:boosting}. The first column indicates whether all the view pairs or just the best pairs (based on $\lambda$ weighting) were used for classification with CNN-1. The second column indicates whether or not the weighting $\lambda$ was used, compared to an equal contribution from each view pair.}
  \label{tab:implementations}
\end{table}

\subsection{Multi-View Recognition}

We then compared our method against two recent competing methods: \emph{ShapeNets} \cite{Wu:etal:CVPR2015} and Multi-View CNN (\emph{MVCNN}) \cite{Su:etal:ICCV2015}, together with a baseline which we call \emph{View Voting}. For ShapeNets, we used code provided by the authors, and we implemented our own version of MVCNN as per the publication details, achieving similar to their quoted results. For the View Voting method, we trained a CNN with 5 convolutional layers and 3 fully-connected layers, similar to CNN-1, to classify views independently based on the image alone. A voting system was then employed to combine the classification outputs of each view. We explored three implementations of our method: using only greyscale images, using only depth images, and then using both image modalities. With the competing methods, ShapeNets was implemented with depth images and MVCNN was implemented greyscale images, as per their original descriptions, and View Voting was implemented with greyscale images.

\begin{table*}
\footnotesize
\centering
  \begin{tabular}{llcccccccc}  
    \toprule
    & & \multicolumn{4}{c}{ModelNet10} & \multicolumn{4}{c}{ModelNet40} \\
    \cmidrule(lr){3-6} \cmidrule(lr){7-10}
    & & 3 views & 6 views & 12 views & \multirow{2}{*}{Average} & 3 views & 6 views & 12 views & \multirow{2}{*}{Average} \\
    Method & Image & (60$^{\circ}$) & (180$^{\circ}$) & (360$^{\circ}$) &  & (60$^{\circ}$) & (180$^{\circ}$) & (360$^{\circ}$) &  \\
    \cmidrule(lr){1-2} \cmidrule(lr){3-6} \cmidrule(lr){7-10}
    View Voting & Greyscale & 87.2 & 89.5 & 90.1 & 88.9 & 85.7 & 87.3 & 88.1 & 87.0 \\
    \cmidrule(lr){1-2} \cmidrule(lr){3-6} \cmidrule(lr){7-10}
    ShapeNets \cite{Wu:etal:CVPR2015} & Depth & 79.2 & 82.5 & 83.1 & 81.6 & 72.0 & 75.7 & 77.4 & 75.0 \\
    \cmidrule(lr){1-2} \cmidrule(lr){3-6} \cmidrule(lr){7-10}
    MVCNN \cite{Su:etal:ICCV2015} & Greyscale & 84.5 & 89.8 & 92.2 & 88.8 & 82.3 & 88.1 & 89.5 & 86.6 \\
    \cmidrule(lr){1-2} \cmidrule(lr){3-6} \cmidrule(lr){7-10}
    \multirow{3}{*}{Ours} & Greyscale & 88.5 & 91.4 & 92.8 & 90.9 & 86.2 & 88.8 & 90.7 & 88.6 \\
    & Depth & 85.2 & 87.6 & 90.0 & 87.6 & 83.0 & 87.0 & 89.9 & 86.6 \\
    & Greyscale + Depth & 88.8 & 91.9 & 93.2 & \textbf{91.3} & 87.0 & 89.5 & 91.1 & \textbf{89.2} \\
    \bottomrule
  \end{tabular}
  \caption{Recognition results over different sequence lengths, from trajectories constrained to rotations about the gravity vector, at an elevation of 30$^{\circ}$. Numbers represent the percentage of correctly-classified objects from the test set.}
  \label{tab:recognition1}
\end{table*}

\begin{table*}
\footnotesize
\centering
  \begin{tabular}{llcccccccc}  
    \toprule
    & & \multicolumn{4}{c}{ModelNet10} & \multicolumn{4}{c}{ModelNet40} \\
    \cmidrule(lr){3-6} \cmidrule(lr){7-10}
    Method & View Selection & 3 views & 6 views & 12 views & Average & 3 views & 6 views & 12 views & Average \\
    \cmidrule(lr){1-2} \cmidrule(lr){3-6} \cmidrule(lr){7-10}
    \multirow{2}{*}{View Voting} & Random & 85.5 & 87.2 & 87.6 & 88.3 & 84.0 & 85.8 & 87.0 & 85.6  \\
    & Straight & 86.1 & 88.8 & 89.9 & 88.3 & 84.3 & 87.0 & 88.1 & 86.5  \\
    \cmidrule(lr){1-2} \cmidrule(lr){3-6} \cmidrule(lr){7-10}
    \multirow{2}{*}{ShapeNets \cite{Wu:etal:CVPR2015}} & Random & 76.0 & 81.6 & 82.2 & 79.9 & 70.0 & 74.4 & 77.2 & 73.9 \\
    & Straight & 77.0 & 81.9 & 82.2 & 80.4 & 70.2 & 74.5 & 77.2 & 74.0 \\
    \cmidrule(lr){1-2} \cmidrule(lr){3-6} \cmidrule(lr){7-10}
    \multirow{2}{*}{Ours} & Random & 86.8 & 87.8 & 91.0 & 88.5 & 86.1 & 88.4 & 90.1 & 88.2 \\
    & Straight & 88.6 & 91.2 & 93.0 & \textbf{90.9} & 86.7 & 89.3 & 91.0 & \textbf{89.0} \\
    \bottomrule
  \end{tabular}
  \caption{Recognition results over different sequence lengths, from unconstrained trajectories. Numbers represent the percentage of correctly-classified objects from the test set.}
  \label{tab:recognition2}
\end{table*}

Table \ref{tab:recognition1} shows recognition results for view sequences at an elevation of 30$^{\circ}$ from the ground plane, constrained to rotations about the gravity vector, as was the experimental setting in \cite{Wu:etal:CVPR2015, Su:etal:ICCV2015}. Our method outperforms all other methods, and the combination of both greyscale and depth images achieves a small boost in performance over single image modality. As was presented in \cite{Su:etal:ICCV2015}, rendering 2D images in a view-based manner achieves much better recognition results than the generative volumetric approach of ShapeNets. However, the capacity of ShapeNets for shape completion provides a strength that view-based methods cannot. The MVCNN method achieves the second-best results for sequences covering 360$^{\circ}$, but for shorter sequences the performance degrades dramatically. This illustrates the unsuitability of their method for arbitrary sequence lengths, owing to testing and training requiring the same sequence length. We note that our method achieves comparable performance to MVCNN with only half the number of views. In our implementation of MVCNN, we trained on the full 360$^{\circ}$ set of images regardless of the length of the testing sequence, although this could likely be improved by training on varying sequence lengths. However, once the constraint of moving only about the gravity vector is removed, this would become intractable.

Table \ref{tab:recognition2} then shows recognition results for arbitrary view sequences, where we exclude MVCNN due to its inability to generalise in this way. Our method here was provided with both greyscale and depth images. We implement each method with two variations of how the next view along the sequence is selected. \emph{Random} chooses the next view randomly from the adjacent views, and \emph{Straight} follows a straight path around the viewing sphere from the beginning to the end of the sequence, similar to the results in \ref{tab:recognition1} except that the sequence direction is randomly selected rather than being constrained around the gravity vector. Again, our method achieves state-of-the-art recognition results, and we note that results are slightly lower than those in Table \ref{tab:recognition1} due to the suitability of viewpoints at an elevation of 30$^{\circ}$ for common household objects in their canonical orientation. Choosing a straight path rather than a random path increases recognition accuracy, due to the tendency of random walks to revisit old viewpoints, or remain within a local region and hence fail to explore the object sufficiently.

\subsection{Active Recognition}

We then evaluated the performance of our NBV and trajectory optimisation extensions to multi-view recognition, and we compared against the NBV method of \emph{ShapeNets} \cite{Wu:etal:CVPR2015}. Our method here was provided with both greyscale and depth images. For both our method and ShapeNets, we implemented two strategies for view selection. \emph{NBV Global} chooses the next-best-view from the entire viewing sphere, and \emph{NBV Adjacent} chooses the next-best-view from the views adjacent to the current view. The \emph{Optimised} implementation of our method is that which incorporates all images along the trajectory rather than just the start and end positions, as in Section \ref{sec:traj_opt}. For all implementations, if the selected viewpoint was one which had already been visited, then the highest-scoring of all the unvisited viewpoints was selected. Table \ref{tab:active} shows that our method achieves the best recognition performance, and whilst the global NBV implementation sees best recognition accuracy, our \emph{Optimised} implementation, which would be used in reality due to its greater practical efficiency, achieves a close second.

\begin{figure}
    \centering
    \includegraphics[width=0.9\linewidth]{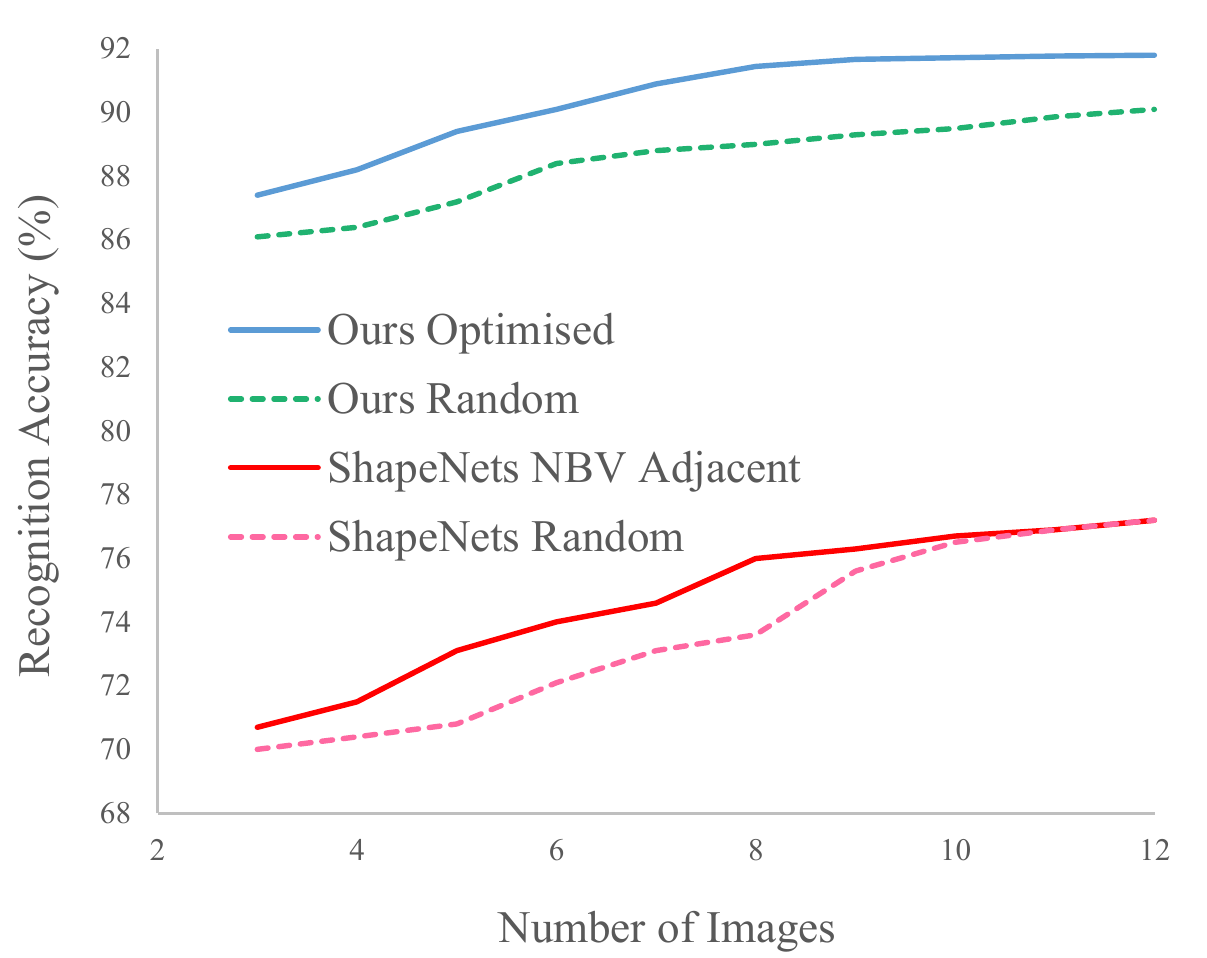}
    \caption{Recognition accuracy for different view sequence lengths on ModelNet40.}
    \label{fig:graph}
\end{figure}

\begin{figure*}
    \centering
    \includegraphics[width=\linewidth]{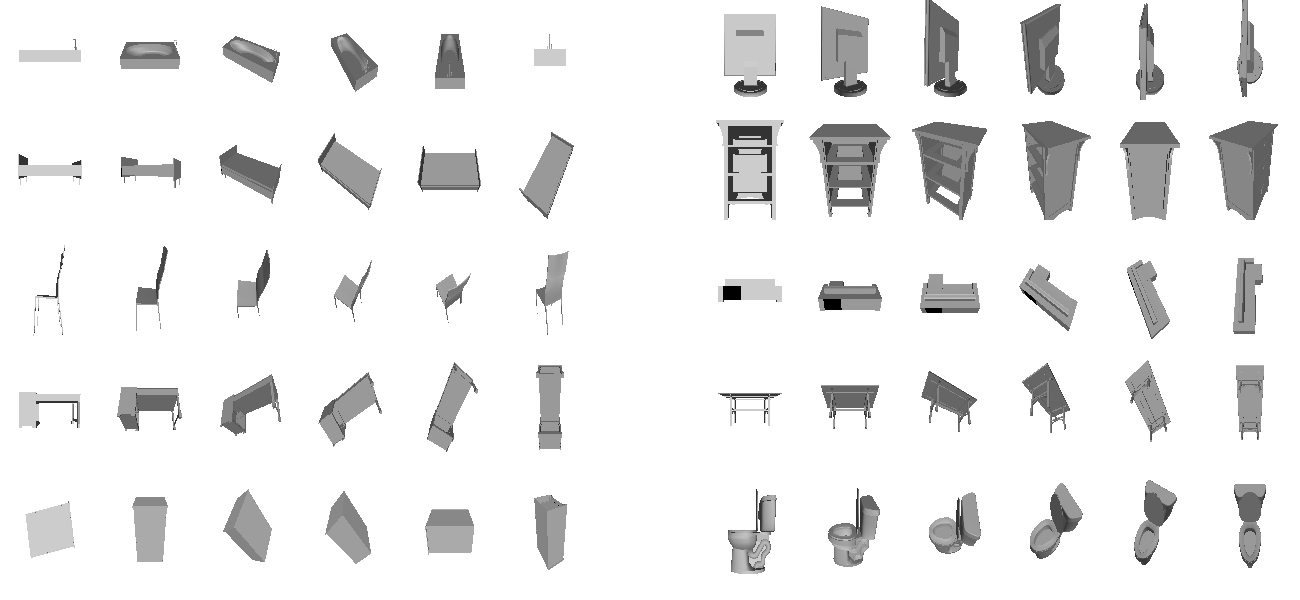}
    \caption{Example greyscale images observed from our optimised trajectories with the ModelNet10 dataset.}
    \label{fig:examples}
\end{figure*}

Figure \ref{fig:graph} plots the recognition accuracy on ModelNet40 against view sequence length for our method and ShapeNets, with each under active and random trajectories and following adjacent viewpoints. Even without trajectory optimisation, our method significantly outperforms ShapeNets, and we see that our trajectory optimisation maintains an advantage over random viewpoint selection, for a range of sequence lengths. Finally, Figure \ref{fig:examples} visualises some image sequences which our method observes under optimised trajectories. We note that the chosen trajectory often passes over the top, or beneath, the object, showing how the constraint of MVCNN to rotations around the gravity vector is sub-optimal.

\begin{table*}
\footnotesize
\centering
  \begin{tabular}{llcccccccc}
    \toprule
    Method & View Selection & \multicolumn{4}{c}{ModelNet10} & \multicolumn{4}{c}{ModelNet40} \\
    \cmidrule(lr){3-6} \cmidrule(lr){7-10}
    & & 3 views & 6 views & 12 views & \multirow{2}{*}{Average} & 3 views & 6 views & 12 views & \multirow{2}{*}{Average} \\
    & & (60\degree) & (180\degree) & (360\degree) &  & (60\degree) & (180\degree) & (360\degree) &  \\
    \cmidrule(lr){1-2} \cmidrule(lr){3-6} \cmidrule(lr){7-10}
    \multirow{2}{*}{ShapeNets \cite{Wu:etal:CVPR2015}} & NBV Global & 79.2 & 82.0 & 82.9 & 81.3 & 71.1 & 75.6 & 77.2 & 74.6 \\
    & NBV Adjacent & 78.7 & 81.0 & 82.4 & 80.7 & 70.7 & 74.2 & 77.2 & 74.0 \\
    \cmidrule(lr){1-2} \cmidrule(lr){3-6} \cmidrule(lr){7-10}
    \multirow{3}{*}{Ours} & NBV Global & 90.4 & 92.8 & 94.0 & \textbf{92.4} & 88.5 & 91.5 & 92.0 & \textbf{90.7} \\
    & NBV Adjacent & 88.8 & 91.6 & 93.5 & 91.3 & 87.2 & 89.5 & 91.4 & 89.4 \\
    & Optimised & 88.9 & 91.9 & 93.9 & 91.6 & 87.4 & 90.1 & 91.8 & 89.8 \\
    \bottomrule
  \end{tabular}
  \caption{Recognition results over different sequence lengths, for unconstrained trajectories, using trajectory optimisation methods. Numbers represent the percentage of correctly-classified objects from the test set.}
  \label{tab:active}
\end{table*}

\section{Conclusions}

In this paper, we have presented a new method for multi-view object recognition over unconstrained camera trajectories, using greyscale images, depth images, or both modalities combined. We have shown that decomposing an image sequence into a set of view pairs enables training in a tractable manner for any trajectory over the viewing sphere of an object. Experiments show that our method outperforms the voxel-based generative ShapeNets method, together with the Multi-View CNN method, and we achieve state-of-the-art recognition on the ModelNet dataset. We have also shown how our pairwise method can extend to next-best-view prediction by learning discriminatively in an end-to-end manner, and this can then be incorporated into a trajectory optimisation scheme to achieve the best camera path for recognition over a given sequence length.

\section{Acknowledgements}
Research presented in this paper has been supported by Dyson Technology Ltd.

\clearpage

{\small 
\clearpage
\bibliographystyle{ieee}
\bibliography{robotvision}
}

\end{document}